\documentclass[conference]{IEEEtran}
\IEEEoverridecommandlockouts
\usepackage{cite}
\usepackage{graphicx}
\usepackage{textcomp}
\usepackage{xcolor}
\usepackage{floatflt}
\usepackage[bookmarks=false]{hyperref}       
\usepackage{url}            
\usepackage{booktabs}       
\usepackage{amsfonts}       
\usepackage{nicefrac}       
\usepackage{microtype}      
\usepackage{amsmath,amssymb}
\usepackage{bbm}
\usepackage{color}
\usepackage{bbm}
\usepackage{multirow}
\usepackage{url}
\usepackage{balance}
\usepackage{multibib}
\usepackage{multicol}
\usepackage{mathtools}
\usepackage{arydshln}
\usepackage{bbm}
\usepackage{balance}
\usepackage{algorithm}
\usepackage[noend]{algpseudocode}
\usepackage{subcaption}
\usepackage{footmisc}
\usepackage{fancyhdr}
\usepackage{fancybox}
\newcommand{\scania}{\texttt{Scania}}
\newcommand{\Fraud}{\texttt{Credit Fraud}}
\newcommand{\pct}{\texttt{PCT Data}}
\newcommand{\cancer}{\texttt{Mammography}}
\newcommand{\eu}{\texttt{PCT Data}}
\newcommand{\webpage}{\texttt{Webpage}}
\newcommand{\protein}{\texttt{Protein Homo}}

\newcommand{\post}{Q}
\newcommand{\posterior}{Q}

\newcommand{\R}{R}
 
\newcommand{\I}{\mathrm{I}}

\newcommand{\Dcal}{{\cal D}}
\newcommand{\Xcal}{{\cal X}}
\newcommand{\Ycal}{{\cal Y}}

\newcommand{\Hcal}{{\cal H}}  

\DeclareMathOperator*{\Esp}{\mathbb{E}} 
\newcommand{\E}[1]{{\displaystyle \Esp_{#1}}}

\newcommand{\sign}{\operatorname{sign}}
\newcommand{\pwr}{^}

\newcommand{\rdt}{\texttt{R-DT}}
\newcommand{\sdt}{\texttt{S-DT}}
\newcommand{\adt}{\texttt{A-DT}}
\newcommand{\rbg}{\texttt{R-BG}}
\newcommand{\sbg}{\texttt{S-BG}}
\newcommand{\abg}{\texttt{A-BG}}
\newcommand{\ee}{\texttt{EE}}
\newcommand{\bb}{\texttt{BB}}
\newcommand{\brf}{\texttt{BRF}}
\newcommand{\our}{\texttt{DAMVI}}

\usepackage{amsthm}
\newtheorem{theorem}{Theorem}
\newtheorem{definition}{Definition}[section]

\def\BibTeX{{\rm B\kern-.05em{\sc i\kern-.025em b}\kern-.08em
    T\kern-.1667em\lower.7ex\hbox{E}\kern-.125emX}}
\begin{document}

\title{Diversity-Aware Weighted Majority Vote Classifier for Imbalanced Data}

\author{\IEEEauthorblockN{ Anil Goyal}
\IEEEauthorblockA{\textit{NEC Laboratories Europe GmbH}\\
	Heidelberg, Germany \\
	anil.goyal@neclab.eu}
\and
\IEEEauthorblockN{ Jihed Khiari}
\IEEEauthorblockA{\textit{NEC Laboratories Europe GmbH}\\
 Heidelberg, Germany \\
jihed.khiari@neclab.eu}
}

\maketitle

\begin{abstract}
In this paper, we propose a diversity-aware ensemble learning based algorithm, referred to as $\our$, to deal with imbalanced binary classification tasks.  Specifically, after learning base classifiers, the algorithm i) increases the weights of positive examples (minority class) which are ``hard'' to classify with uniformly weighted base classifiers; and ii) then learns weights over base classifiers by optimizing the PAC-Bayesian C-Bound that takes into account the accuracy and diversity between the classifiers. 
We show efficiency of the proposed approach with respect to state-of-art models on predictive maintenance task, credit card fraud detection,  webpage classification and medical applications. 
\end{abstract}


\section{Introduction}
\label{sec:intro}
Most machine learning algorithms assume that underlying class distribution (i.e. percentage of examples belonging to each class) is balanced. 
However, in many real-world applications (e.g. anomaly detection, medical diagnosis, predictive maintenance, driver behavior detection or detection of oil spills),  the number of examples from negative class (majority class) significantly outnumbers the number of positive class (minority class or class of interest) examples.
In such situations, the traditional machine learning algorithms tend to have bias towards the majority class. 
This problem of machine learning is known as imbalanced learning or learning from imbalanced data \cite{he2009learning}. \\[2mm]
\noindent \textbf{Related Work.} 
In the literature, many studies have been conducted to address the problem of imbalanced learning. 
Most of the proposed approaches can be categorized into $3$ groups depending on the way they deal with class imbalance. 
Data level approaches\cite{batista2004study,chawla2002smote,he2008adasyn,liu2007generative,liu2008exploratory} focus on balancing the input data distribution in order to reduce the effect of class imbalance during the learning process. 
The  algorithm level\cite{fan1999adacost,galar2011review,he2009learning,zadrozny2003cost,wu2005kba} approaches focus on developing or modifying the existing algorithms to handle imbalanced datasets by giving more significance to positive examples.
Finally, the cost-sensitive approaches\cite{elkan2001foundations,thai2010cost,ling2008cost,yuan2018regularized} deals with class imbalance by incorporating different classification costs for each class.

Among these approaches, a group of techniques make use of ensembles of classifiers.
Ensemble learning \cite{Dietterich00,re2012ensemble},  aims at combining a set of classifiers in order to build a more efficient classifier than each of the individual classifier alone. 
This strategy has shown to be effective in large number of applications\cite{oza2008classifier,zhang2012ensemble}.
While dealing with imbalanced data, one of the main advantages of ensemble learning approaches is that they are versatile to the choice of base learning algorithm.
Many ensemble learning based approaches have been proposed to deal with imbalanced datasets, including but not limited to EasyEnsemble\cite{liu2008exploratory}, SMOTEBagging\cite{wang2009diversity}, Balanced Random Forest\cite{khoshgoftaar2007empirical} or SMOTEBoost\cite{chawla2003smoteboost}. 
In the ensemble learning literature, it is well known that controlling the trade-off between accuracy and diversity among classifiers  plays a key role while learning a combination of classifiers \cite{brown10,Kuncheva}.
Moreover, \textit{D{í}ez-Pastor et. al.}\cite{Pastor15} and \textit{Yao et. al.}\cite{wang2009diversity} showed that approaches that control the diversity among classifiers improves the performance of imbalanced classification tasks.
With this in mind, our objective is to design an algorithm for imbalanced datasets which explicitly controls this trade-off between accuracy and diversity among classifiers. \\[2mm]
\noindent \textbf{Contribution.}
In this work, we propose an ensemble method  that outputs a \textbf{D}iversity-\textbf{A}ware weighted \textbf{M}ajority \textbf{V}ote over previously learned base classifiers for \textbf{I}mbalanced datasets (referred to as $\our$).
In order to learn weights over the base classifiers, we minimize the upper bound on the error of the majority vote, using PAC-Bayesian $C$-Bound \cite{Lacasse06,GermainLLMR15}, which allows us to control the trade-off between accuracy and diversity. 
Concretely, after learning base classifiers for different bootstrapped samples of input data, the algorithm i) increases the weights of positive examples (minority class) which are ``hard'' to classify with uniformly weighted base classifiers; and ii) then learns weights over base classifiers by optimizing the C-Bound (with focus on ``hard''positive examples).
The key benefits of our approach are that it does not make any prior assumption on underlying data distribution and it is independent of base learning algorithm.
To show the potential of our algorithm, we empirically evaluate our approach on predictive maintenance task, credit card fraud detection,  webpage classification and medical applications. 
From our experiments, we show that $\our$ is more ``consistent''and ``stable'' compared to state-of-art methods both in terms of F1-measure and Average Precision (AP), in case when we have high imbalance in class distribution ($<4\%$ of Imbalance Ratio). This is due to the fact that our method is able to explicitly control the trade-off between accuracy and diversity among classifiers on hard positive examples.
\\[2mm]
\noindent \textbf{Paper Organization.}
In the next section, we present general notations and setting for our algorithm. In Section \ref{sec:algo}, we derive our algorithm $\our$ for imbalance datasets. Before concluding in Section \ref{sec:conclusion}, we present obtained experimental results using our approach in Section \ref{sec:exp}.



\section{Notations and Setting}
\label{sec:notations}
In this work, we consider a binary classification task where the examples are drawn from a fixed yet unknown distribution $\Dcal$ over $\Xcal \times \Ycal$, where $\Xcal \subseteq \mathbb{R}^d $ is the $d$-dimensional input space and $\Ycal = \{-1,+1\}$ the label/output space. 
Typically, in case of learning with imbalanced data, the percentage of examples belonging to one class is significantly  smaller than the another class.
In our case, we assume that examples belonging to positive class are in minority. 
A learning algorithm is provided with training sample of $n$ examples denoted by $S = \{ (x_i, y_i) \}_{i=1} \pwr n$, that is assumed to be independently and identically distributed (\textit{i.i.d.}) according to $\Dcal$.
We further assume that we have a set of classifiers $\Hcal$ from $\Xcal$ to $\Ycal$.
Given $S$, our objective is to learn a weight distribution $\post$  over $\Hcal$ that leads to a well performing weighted majority vote ($B_{\post}$), such that $B_{\post}$ 
\begin{align}
\label{eq:MajorityVote}
B_{\post}(x) = \sign \left[ \E{h \sim \post} h(x) \right]
\end{align}
has the smallest possible generalization error on $\Dcal$ which is highly imbalanced in terms of class distribution.
In other words, our goal is to learn $Q$ over $\Hcal$ such that it minimizes the true risk $\R_{\Dcal}(B_{\post})$ of $B_{\post}(x)$:
\begin{align}
\label{eq:TrueRisk}
R_{\Dcal}(B_{\post}) = \E{(x,y) \sim \Dcal} \I \left[ B_{\post}(x) \neq y \right]
\end{align}
where, $\I \left[ p \right]$ is equal to $1$ if the predicate $p$ is true, and $0$ otherwise. 
An important behavior of the above risk on $\post$-weighted majority vote $B_{\post}$ is that it is closely related to the Gibbs risk $R_{\Dcal}(G_{Q})$ which is defined as  the expectation of the individual risks of each classifier that appears in the majority vote. Formally, we can define Gibbs risk as follows:
\begin{align*}
& R_{\Dcal}(G_{\post}) = \E{(x,y) \sim \Dcal}\ \E{h \sim Q} \I[h(x) \neq y] .   
\end{align*}
In fact, if $B_Q$ misclassifies $x \in \Xcal$, then at least half of the classifiers (under measure $Q$) makes a prediction error on $x$. Therefore. we have
\begin{align*}
    R_{\Dcal}(B_Q) \leq 2 R_{\Dcal}(G_Q)
\end{align*}
Thus, an upper bound on $R_{\Dcal}(G_Q)$ gives rise to an upper bound on $R_{\Dcal}(B_Q)$.
There exist other tighter relations\cite{Lacasse06,GermainLLMR15,LangfordS02}, such as PAC-Bayesian $C$-Bound\cite{Lacasse06} that involves the \textit{expected disagreement} $d_{\Dcal} (Q)$ between pair of classifiers, and that can be expressed as follows (when $R_{\Dcal}(G_{\post}) \leq \frac{1}{2}$):
\begin{align}
\label{eq:Cbound}
& R_{\Dcal}(B_{Q}) \leq 1 - \frac{\left(1- 2 R_{\Dcal}(G_{Q}) \right) \pwr 2}{1-2d_{\Dcal} (\post)}\\
\nonumber & \text{where } d_{\Dcal} (Q)  = \E{(x,y) \sim \Dcal}\ \E{h \sim Q}\ \E{h' \sim Q} \I[h(x) \neq h'(x)] 
\end{align}
We provide the proof of above $C$-Bound in Appendix \ref{sec:proof_c_bound}. 
The expected disagreement $d_{\Dcal} (Q)$  measures the diversity/disagreement among classifiers.
It is worth noting that from imbalanced data classification standpoint where the notion of diversity among classifiers is known to be important (\cite{Pastor15,wang2009diversity}), Equation \ref{eq:Cbound} directly captures the trade-off between the accuracy and the diversity among classifiers. 
Therefore, in this work, we propose a new algorithm (presented in next Section \ref{sec:algo}) for imbalanced learning which directly exploits PAC-Bayesian $C$-Bound in order to learn a weighted majority vote classifier.  
Note that, the PAC-Bayesian $C$-Bound has been shown to be an effective approach to learn a weighted majority vote over a set of classifiers in many applications, e.g. multimedia analysis\cite{morvant2014majority} and multiview learning\cite{goyal2017pac,GOYAL201981}.

\section{Learning a Majority Vote for Imbalanced Data}
\label{sec:algo}
Our objective is to learn weights over a set of classifiers that leads to a well-performing weighted majority vote (given by Equation \ref{eq:MajorityVote}) to deal with imbalanced datasets.
It has been shown that controlling the trade-off between accuracy and diversity between the set of classifiers plays an important role for imbalanced classification problems \cite{Pastor15,wang2009diversity}.
Therefore, we utilize PAC-Bayesian $C$-Bound (given by Equation \ref{eq:Cbound}) which explicitly controls this trade-off in order to derive a diversity-aware ensemble learning based algorithm (referred as $\our$, see Algorithm \ref{algorithm}) for binary imbalanced classification tasks. 
\begin{algorithm}[t]
\caption{$\our$}\label{algorithm}
\textbf{Input:}  Training set $S = \{({x}_i, y_i), \dots , ({x}_n, y_n)\}$, where $x_i \subseteq \text{\hspace{1cm}} \mathbb{R}^d $ and $y_i \in \{-1,+1\}$, \\ 
\text{\hspace{1cm}}Number of classifiers $K$ \\ 
\text{\hspace{1cm}}Base Learning algorithm $A$. \\
\textbf{Initialize:} Empty set of classifiers $\Hcal = \{\phi\}$. \\
\text{\hspace{1.5cm}} $\forall \ x_i \in S, \Dcal(x_i) \gets \frac{1}{n}$
\begin{algorithmic}[1]
\For{k = 1, $\dots$, K}
		\State Generate a bootstrap sample $S(k)$ from $S$
		\State Learn a classifier $h_k$ using the base learning algorithm $A$
		\State Update  $\Hcal = \Hcal \cup \{h_k\}$
\EndFor	
\For{$h_k \in \Hcal$}
    \State $\post(h_k) \gets \frac{1}{K}$ 
\EndFor
\State {Update} the distribution $\Dcal$ over the learning sample $S$
\begin{align*}
&\!\!\!\!\!\!\forall x_i \! \in \! S, \Dcal(x_i) \! = \! 
\begin{cases}
\frac{\Dcal(x_i) \exp \left( -y_i \sum_{k=1}\pwr K Q(h_k) h_k(x_i) \right)}{Z}, &\!\!\!\! \!   \text{if } y_i \! = \! 1  \\
\frac{\Dcal(x_i)}{Z}, & \!\! \!\! \!\!\!\!\! \text{if } y_i \! = \! -1
\end{cases} \\
&\!\!\!\!\!\! \text{where, } Z = \sum_{j=1} \pwr n \Dcal(x_j) \text{ is a normalization factor.}
\end{align*}
\State \textbf{Optimize} the $C$-Bound to learn weights over classifiers
\begin{align*}
&\!\!\!\!\!\! \underset{\post}{\max}  \quad\!\!\!  \left(  \left[ 1 - 2\ {\sum_{i=1}\pwr n \  \Dcal(x_i) \sum_{k=1}\pwr K \post(h_k) \I\left[h_k(x_i) \neq y_i\right] } \right] \pwr 2 \bigg/ \right. \\
& \!\!\!\!\!\!\!\!\!\! \left. \left[ 1 \!- \!2 \sum_{i=1}\pwr n \! \Dcal(x_i) \sum_{k=1} \pwr K \! \sum_{k'=1} \pwr K \! \post(h_k) \post(h_{k'})  \I \left[h_k(x_i) \neq h_{k'}(x_i)\right] \right] \! \right) \\
&\!\!\!\!\!\! \text{s.t.} \qquad \  \sum_{k=1} \pwr K \post(h_k) = 1, \  \post(h_k) \geq 0 \ \ \forall k \in \{1, \dots, K\} 
\end{align*} 

\State \textbf{Return:} Set of classifiers $\Hcal$ and learned weights over classifiers i.e. $Q$. Such that, for any input example $x$, final weighted majority is defined as:
\begin{align*}
B_{\post}(x) = \sign \left( \sum_{k=1} \pwr K \post(h_k) h_k(x)  \right)
\end{align*}
\end{algorithmic}
\end{algorithm}

For a given training set $S = \{(x_1,y_1), \dots, (x_n,y_n)\} \in (\mathbb{R}^d \times \{-1,+1\}) \pwr n$ of size $n$; 
$\our$ (Algorithm \ref{algorithm}) trains a set of base classifiers $\Hcal = \{h_1, \dots, h_K\}$ (using a base learning algorithm $A$) corresponding to $K$ bootstrapped samples (Step $1$ to $4$) 
\footnote{Our algorithm is not limited to base learners learnt using bootstrapped samples. It is applicable to any set of base learners.}.
Then, we propose to update the weights of those training examples which belong to the minority class (in our case , $y_i = 1$) as follows (Step $7$): 
\begin{align*}
&\forall x_i \in S, \  \Dcal(x_i) =  
\begin{cases}
\frac{\Dcal(x_i) \exp \left( -y_i \sum_{k=1}\pwr K Q(h_k) h_k(x_i) \right)}{Z}, &\!\!\!\! \!   \text{if } y_i \! = \! 1  \\
\frac{\Dcal(x_i)}{Z}, & \!\! \!\! \!\!\!\!\! \text{if } y_i \! = \! -1
\end{cases} \\
& \text{where, } Z = \sum_{j=1} \pwr n \Dcal(x_j) \text{ is a normalization factor.}
\end{align*}
In Step $7$, the weights of misclassified (resp. correctly classified) positive examples according to the uniformly weighted majority vote classifier increase (resp. decrease). 
Note that, here we update the weights over the learning sample $S$ just once by focusing only on positive examples. Whereas, boosting algorithms\cite{Freund95} (e.g. Adaboost\cite{adaboost}) repeatedly learn a ``weak'' classifier using a learning algorithm with different probability distribution over $S$.
Intuitively, this step increases the weights of those positive examples which are ``hard''to classify with the uniformly weighted classifier ensemble.
This step allows us to focus on ``hard'' positive examples while learning weights over the base classifiers.

Then, we propose to learn the weights over the classifiers by optimizing the $C$-Bound on weighted training sample $S$, given by Equation \ref{eq:Cbound} (Step $8$), which can be represented by the following constraint optimization problem: 
\begin{align*}
& \underset{\post}{\max}  \quad  \left(  \left[ 1 - 2\ {\sum_{i=1}\pwr n \  \Dcal(x_i) \sum_{k=1}\pwr K \post(h_k) \I\left[h_k(x_i) \neq y_i\right] } \right] \pwr 2 \bigg/ \right. \\
& \!\! \left. \left[ 1 \!- \!2 \sum_{i=1}\pwr n \! \Dcal(x_i) \sum_{k=1} \pwr K \! \sum_{k'=1} \pwr K \! \post(h_k) \post(h_{k'})  \I \left[h_k(x_i) \neq h_{k'}(x_i)\right] \right] \right) \\
& \text{s.t.} \qquad \  \sum_{k=1} \pwr K \post(h_k) = 1, \  \post(h_k) \geq 0 \ \ \forall k \in \{1, \dots, K\} 
\end{align*} 
Intuitively, on ``hard'' positive examples, the $C$-Bound tries to diversify the classifiers and at the same time controls the classification error of the classifiers which is a key element for imbalanced datasets\cite{Pastor15,wang2009diversity}. 
As above optimization problem is constrained nonlinear problem, therefore we use 
Sequential Quadratic Programming \cite{nocedal2006numerical} algorithm which uses the quasi-Newton method to find maxima of above optimization problem. 

Finally, the learned weights over the classifiers leads to a well-performing majority vote, given by Equation \ref{eq:MajorityVote}, tailored for imbalanced classification tasks. 
For any input example $x$, the final learned weighted majority vote is given as follows:
\begin{align*}
B_{\post}(x) = \sign \left( \sum_{k=1} \pwr K \post(h_k) h_k(x)  \right)
\end{align*}

\section{Experiments}
\label{sec:exp}
\setlength{\tabcolsep}{4pt}
In this section, we present an empirical study to show the performance of our algorithm $\our$ on following datasets.

\begin{table}[h]
	\begin{center}
		\caption{Summary of Datasets: Number of attributes, Number of examples and the Imbalance Ratio (IR) i.e. percentage of positive examples (minority class).}
		\begin{tabular}{@{\extracolsep{\fill}} p{2cm} p{1.5cm} p{1.5cm} p{1cm} }
				\toprule
				\label{table:datasets}
				&\rotatebox[origin=l]{0}{\#Attributes}
				&\rotatebox[origin=l]{0}{\#Examples}
				&\rotatebox[origin=l]{0}{IR}
				\\	
				\hline
				$\webpage$&300&34780&3.03\\
				$\cancer$&6&11183&2.32\\
				$\scania$&170&60000&1.67\\
				$\protein$&74&145751&0.9\\
				$\Fraud$&30&284807&0.17\\
				$\eu$&17&816099&0.02\\
				\hline
		\end{tabular}
	\end{center}		
\end{table}

\subsection{Datasets}
We have validated $\our$ on  $6$ datasets belonging to predictive maintenance task, credit card fraud detection,  webpage classification and medical applications. 
A description of these datasets is presented in Table \ref{table:datasets}.
\begin{itemize}
    \item \textbf{Predictive maintenance} relies on equipment data (telemetry data) and historical maintenance data to track the performance of equipment in order to predict possible failures in advance. We considered real-world  $\scania$ dataset\cite{Dua:2019}\footnote{\url{https://archive.ics.uci.edu/ml/datasets/APS+Failure+at+Scania+Trucks}} which is openly available and collected from heavy Scania trucks in everyday usage. The positive class (minority class) corresponds to failures of specific component of the Air Pressure System (APS) and negative class corresponds to failures of components not related to the APS system. The $\pct$  consists of equipment data (sensor data) and maintenance data from trucks operating at Piraeus Container Terminal (PCT) in Athens, Greece. The positive class (minority class) corresponds to truck failures and the negative class corresponds to normally functioning trucks. This dataset is proprietary and was obtained thanks to a research collaboration. 
    \item \noindent \textbf{Credit Card Fraud Detection} composed of credit card transactions where positive class (minority class) examples are fraudulent transactions and negative class examples are non-fraudulent. The $\Fraud$ dataset\cite{dal2015calibrating}\footnote{\url{https://www.kaggle.com/mlg-ulb/creditcardfraud}} is an openly available real-world dataset  consisting of credit card transactions occurred during two days in September, 2013. This dataset was collected and analyzed during a research collaboration between Worldline and ULB (Université Libre de Bruxelles).
    \item \textbf{Medical Datasets: } We considered $2$ openly available datasets related to medical applications: $\cancer$ and $\protein$\footnote{\label{footnote:imblearn}\url{https://imbalanced-learn.readthedocs.io/en/stable/generated/imblearn.datasets.fetch_datasets.html}}. The $\cancer$ dataset\cite{JMLR:v18:16-365} composed of  results from an eponymous breast screening method. The positive class (minority class) corresponds to a malignant mass and the negative class corresponds to a benign mass. The $\protein$ dataset\cite{caruana2004kdd} is an openly available dataset from 2004 KDD-Cup competition\footref{footnote:imblearn}. It is a protein homology prediction task where homologous (\textit{resp.} non-homologous) sequences correspond to the positive (\textit{resp.} negative) class. 
    \item $\webpage$ \cite{JMLR:v18:16-365} is an openly available text classification dataset\footref{footnote:imblearn} where the objective is to identify whether a webpage belongs to a particular category (positive class) or not.
\end{itemize}

\subsection{Experimental Protocol}
To study the performance of $\our$, we considered following $9$ baseline approaches\cite{galar2011review}: 
\begin{itemize}
	\item \textbf{Random Oversampling $+$ Decision Tree($\rdt$):} This approach first balances the class distribution by randomly replicating minority class examples. Then, we learn a decision tree classifier on oversampled data.
	\item \textbf{SMOTE $+$ Decision Tree ($\sdt$):} This approach first oversamples the minority class examples using Synthetic Minority Over Sampling Technique (SMOTE) \cite{chawla2002smote} algorithm. 
SMOTE oversamples the minority class examples by interpolating between several minority class examples that lie together. 
After oversampling, we learn a decision tree classifier.
	
	\item \textbf{ADASYN $+$ Decision Tree ($\adt$):} This approach first oversamples the minority class examples using Adaptive Synthetic (ADASYN) sampling algorithm \cite{he2008adasyn}. 
ADASYN computes a weight distribution over minority class examples to synthetically generate data for minority class examples that are harder to learn.
After oversampling, we learn a decision tree classifier.
	
	\item \textbf{ROSBagging ($\rbg$)}\cite{galar2011review}: This approach first oversamples the minority class examples by following Random Oversampling (ROS) approach. Then, we learn an ensemble of decision tree classifiers  on bootstrapped samples of oversampled data. 	
	\item \textbf{SMOTEBagging ($\sbg$)}\cite{chawla2003smoteboost}:This approach first oversamples the minority class examples following SMOTE algorithm. Then, we learn an ensemble of decision tree classifiers  on bootstrapped samples of oversampled data. 	  
	
	\item \textbf{ADASYNBagging ($\abg$)}:This approach first oversamples the minority class examples following ADASYN algorithm. Then, we learn an ensemble of decision tree classifiers  on bootstrapped samples of oversampled data. 
	
	\item \textbf{Balanced Bagging ($\bb$)}\cite{galar2011review}: This approach balances the dataset using random undersampling. Then, an ensemble of decision tree classifiers  are learnt on bootstrapped samples of oversampled data. 	
	\item \textbf{Balanced Random Forest ($\brf$)}\cite{khoshgoftaar2007empirical}: This approach learns an ensemble of classification trees from balanced bootstrapped samples of original input data.
	
	\item \textbf{Easy Ensemble ($\ee$)}\cite{liu2008exploratory}: This approach learns an ensemble of AdaBoost learners trained on different balanced bootstrap samples.
\end{itemize}
For all oversampling based approaches ($\rdt , \sdt , \adt, \rbg, \sbg, \abg$), we used the ROS, SMOTE and ADASYN implementations of \texttt{imbalanced-learn} python package \cite{JMLR:v18:16-365} to synthetically generate new minority class examples such that the number of minority class examples is equal to the number of majority class examples. 
For SMOTE and ADASYN, we considered $5$ nearest neighbours to generate synthetic examples.

For $\ee, \bb$ and $\brf$, we used implementations of \texttt{imbalanced-learn} python package with number of base learners equals to $100$. 
For our approach $\our$\footnote{$\our$ codes are available at \url{https://github.com/goyalanil/DAMVI}} and baselines $\rbg, \sbg, \abg$, we fix the number of decision tree classifiers to $100$ and size of bootstrapped sample to $20\%$ of the size of original training data. 
For our approach $\our$, we learn the weights over base classifiers by optimizing $C$-Bound on weighted training sample $S$.
For solving the constrained optimization problem, we used Sequential Least SQuares Programming (SLSQP) implementation of scikit-learn \cite{scikit-learn} (that we also used to learn the decision tree classifiers) with uniform initialization of weights over the base classifiers.
For all the experiments, we reserved $30\%$ of data for testing and the remaining for training.
Experiments are repeated $5$ times by each time splitting the training and the test sets at random over the
initial datasets.\\

\noindent \textbf{Evaluation Metrics:} Under the imbalanced learning scenario, the conventional evaluation metrics such as accuracy are unable to adequately represent the model's performance on the minority class examples which is typically the class of interest\cite{he2009learning}.
Therefore, we evaluate  the models based on two metrics: F1-score and Average Precision (AP), which are known to be relevant for imbalanced classification problems\cite{davis2006relationship,galar2011review,he2009learning}. 
F1-score is defined as harmonic mean of precision and recall.
Whereas, Average Precision (AP) is the area under the precision-recall curve and it has been shown that AP, in case of highly imbalanced datasets, is more informative than AUC ROC \cite{davis2006relationship}.

\begin{table*}[ht]
	\caption{F1-score for different approaches averaged over $5$ random sets. Along each column, the best result is in bold, and second one in italic. $\pwr \downarrow$ indicates that a result is statistically significantly worse than $\our$, according to Wilcoxon rank sum test\cite{StatsRankMethods} with $p<0.05$}
	\begin{center}
			\begin{tabular}{l | c c c c c c }
				\hline
				       & $\webpage$                               & $\cancer$                                & $\scania$                                                           & $\protein$                      & $\Fraud$                  & $\pct$               \\ \hline
				$\sdt$ & .5062$\pm$.028$\pwr \downarrow$          & .5198$\pm$.009$\pwr \downarrow$          & .5848$\pm$.011$\pwr \downarrow$            & .5278$\pm$.005$\pwr \downarrow$ & .5610$\pm$.016$\pwr \downarrow$  &.8793$\pm$.011$\pwr \downarrow$        \\
				$\rdt$ & .4705$\pm$.021$\pwr \downarrow$          & .6053$\pm$.043$\pwr \downarrow$          & .6256$\pm$.019$\pwr \downarrow$                 & .7290$\pm$.017$\pwr \downarrow$ & .7556$\pm$.013$\pwr \downarrow$      & \textit{.9715}$\pm$.002$\pwr \downarrow$      \\
				$\adt$ & .4693$\pm$.019$\pwr \downarrow$          & .4978$\pm$.034$\pwr \downarrow$          & .5807$\pm$.020$\pwr \downarrow$            & .5259$\pm$.019$\pwr \downarrow$ & .5653$\pm$.027$\pwr \downarrow$     & .8830$\pm$.009$\pwr \downarrow$     \\
				$\rbg$ & .4620$\pm$.016$\pwr \downarrow$          & \textit{.6145}$\pm$.026$\pwr \downarrow$ & \textit{.6845}$\pm$.014$\pwr \downarrow$           & \textit{.7849}$\pm$.021$\pwr \downarrow$         & \textit{.7703}$\pm$.020$\pwr \downarrow$ & {.9691}$\pm$.001$\pwr \downarrow$\\
				$\sbg$ & \textit{.6134}$\pm$.017$\pwr \downarrow$ & .5391$\pm$.017$\pwr \downarrow$          & .6493$\pm$.009$\pwr \downarrow$            & .6771$\pm$.009$\pwr \downarrow$ & .6839$\pm$.024$\pwr \downarrow$  & .9430$\pm$.006$\pwr \downarrow$        \\
				$\abg$ & .4804$\pm$.021$\pwr \downarrow$          & .5169$\pm$.011$\pwr \downarrow$          & .6269$\pm$.007$\pwr \downarrow$            & .6346$\pm$.013$\pwr \downarrow$ & .6819$\pm$.030$\pwr \downarrow$     & .9312$\pm$.004$\pwr \downarrow$     \\
				$\bb$  & .3445$\pm$.001$\pwr \downarrow$          & .4465$\pm$.030$\pwr \downarrow$          & .4317$\pm$.005$\pwr \downarrow$           & .4275$\pm$.008$\pwr \downarrow$ & .1376$\pm$.006$\pwr \downarrow$   & .8014$\pm$.006$\pwr \downarrow$        \\
				$\brf$ & .4098$\pm$.010$\pwr \downarrow$          & .3659$\pm$.014$\pwr \downarrow$          & .3822$\pm$.004$\pwr \downarrow$            & .4027$\pm$.009$\pwr \downarrow$ & .1255$\pm$.016$\pwr \downarrow$     & .2943$\pm$.007$\pwr \downarrow$     \\
				$\ee$  & .4678$\pm$.011$\pwr \downarrow$          & .2534$\pm$.002$\pwr \downarrow$          & .4096$\pm$.006$\pwr \downarrow$          & .3350$\pm$.003$\pwr \downarrow$ & .0922$\pm$.007$\pwr \downarrow$       & .0881$\pm$.001$\pwr \downarrow$    \\ \hline
				$\our$ & \textbf{.7996}$\pm$.011                  & \textbf{.6661}$\pm$.023                  & \textbf{.7289}$\pm$.011                                    & \textbf{.8067}$\pm$.009         & \textbf{.8495}$\pm$.019      & \textbf{.9816}$\pm$.001             \\ \hline
			\end{tabular}
	\end{center}
	\label{table:f1_results}
\end{table*}
\begin{table*}[ht]
	\caption{Average Precision (AP) for different approaches averaged over $5$ random sets. Along each column, the best result is in bold, and second one in italic. $\pwr \downarrow$ indicates that a result is statistically significantly worse than $\our$, according to Wilcoxon rank sum test\cite{StatsRankMethods} with $p<0.05$}
	\begin{center}
			\begin{tabular}{l | c c c c c c }
				\hline
				       & $\webpage$                               & $\cancer$                       & $\scania$                                                        & $\protein$                      & $\Fraud$             & $\pct$             \\ \hline
				$\sdt$ & .2794$\pm$.023$\pwr \downarrow$          & .2919$\pm$.010$\pwr \downarrow$ & .3526$\pm$.014$\pwr \downarrow$           & .3153$\pm$.005$\pwr \downarrow$ & .3482$\pm$.016$\pwr \downarrow$ & .7785$\pm$.001$\pwr \downarrow$ \\
				$\rdt$ & .3008$\pm$.016$\pwr \downarrow$          & .3811$\pm$.054$\pwr \downarrow$ & .3994$\pm$.024$\pwr \downarrow$          & .5347$\pm$.025$\pwr \downarrow$ & .5728$\pm$.020$\pwr \downarrow$  & .9447$\pm$.005$\pwr \downarrow$ \\
				$\adt$ & .2481$\pm$.014$\pwr \downarrow$          & .2740$\pm$.034$\pwr \downarrow$ & .3483$\pm$.023$\pwr \downarrow$          & .3112$\pm$.019$\pwr \downarrow$ & .3516$\pm$.030$\pwr \downarrow$  & .7851$\pm$.016$\pwr \downarrow$ \\
				$\rbg$ & .4944$\pm$.010$\pwr \downarrow$          & \textit{.7011}$\pm$.021         & \textit{.8097}$\pm$.016$\pwr \downarrow$         & .8495$\pm$.016                  & .8120$\pm$.030$\pwr \downarrow$ & \textit{.9875}$\pm$.001 \\
				$\sbg$ & .6219$\pm$.028$\pwr \downarrow$          & .6971$\pm$.025$\pwr \downarrow$                  & .7275$\pm$.019$\pwr \downarrow$           & .8424$\pm$.013                  & .8135$\pm$.027$\pwr \downarrow$     & .9863$\pm$.001$\pwr \downarrow$             \\
				$\abg$ & .4400$\pm$.024$\pwr \downarrow$          & .6261$\pm$.036$\pwr \downarrow$ & .6712$\pm$.018$\pwr \downarrow$           & .8276$\pm$.016                  & \textit{.8137}$\pm$.035$\pwr \downarrow$    & .9847$\pm$.005$\pwr \downarrow$     \\
				$\bb$  & .6302$\pm$.034$\pwr \downarrow$          & .6644$\pm$.037$\pwr \downarrow$ & .6745$\pm$.024$\pwr \downarrow$         & .8359$\pm$.018                  & .7516$\pm$.048$\pwr \downarrow$  & .9849$\pm$.001$\pwr \downarrow$  \\
				$\brf$ & .6930$\pm$.022$\pwr \downarrow$          & .6782$\pm$.023                  & .6877$\pm$.016$\pwr \downarrow$          & \textit{.8549}$\pm$.014         & .7615$\pm$.047$\pwr \downarrow$ & .6976$\pm$.010$\pwr \downarrow$ \\
				$\ee$  & \textit{.6969}$\pm$.038$\pwr \downarrow$ & .5967$\pm$.043$\pwr \downarrow$ & .7558$\pm$.014$\pwr \downarrow$           & \textbf{.8561}$\pm$.012         & .7672$\pm$.025$\pwr \downarrow$ & .0790$\pm$.001$\pwr \downarrow$ \\ \hline
				$\our$ & \textbf{.8331}$\pm$.013                  & \textbf{.7142}$\pm$.039         & \textbf{.8335}$\pm$.007                         & .8267$\pm$.013                  & \textbf{.8373}$\pm$.027    & \textbf{.9976}$\pm$.001       \\ \hline
			\end{tabular}	
	\end{center}
	\label{table:ap_results}
\end{table*}
\subsection{Results}
Firstly, we report the comparison of our algorithm $\our$ with all the considered baselines in Table \ref{table:f1_results} (for F1-score) and Table \ref{table:ap_results} (for Average Precision).
As shown in Tables \ref{table:f1_results} and \ref{table:ap_results}, our proposed algorithm $\our$ performs best compared to baseline approaches for all datasets in terms of F1-score and for $5$ out of $6$ datasets in terms of Average Precision. 
Moreover, on $\pct$  (where we have lowest imbalance ratio i.e. $0.02$), we perform significantly better than the baselines.
According to Wilcoxon rank sum test\cite{StatsRankMethods}, in most of cases, we are significantly better than the baselines with $p<0.05$. 
We can also remark that $\our$ is more ``stable'' than $\rbg$ (in general, second best approach) according to standard deviation values. 
Note that $\rbg, \sbg, \abg, \ee, \bb$ and $\brf$ are able to create a diverse set of base classifiers on bootstrapped samples of input data. However, these approaches don't focus on learning the weights over the base classifiers tailored for imbalanced datasets. Whereas, $\our$  explicitly learns the weights by controlling the trade-off between the accuracy and the diversity among base classifiers  by minimizing PAC-Bayesian $C$-Bound (with focus on ``hard'' positive examples). 
Our results provide evidence that learning a diversity-aware weighted majority vote classifier is an effective way  to deal with imbalanced datasets. 


\begin{figure}[ht] 
	\begin{subfigure}{1.0\textwidth}
		\hspace{-15pt} \includegraphics[scale=0.5]{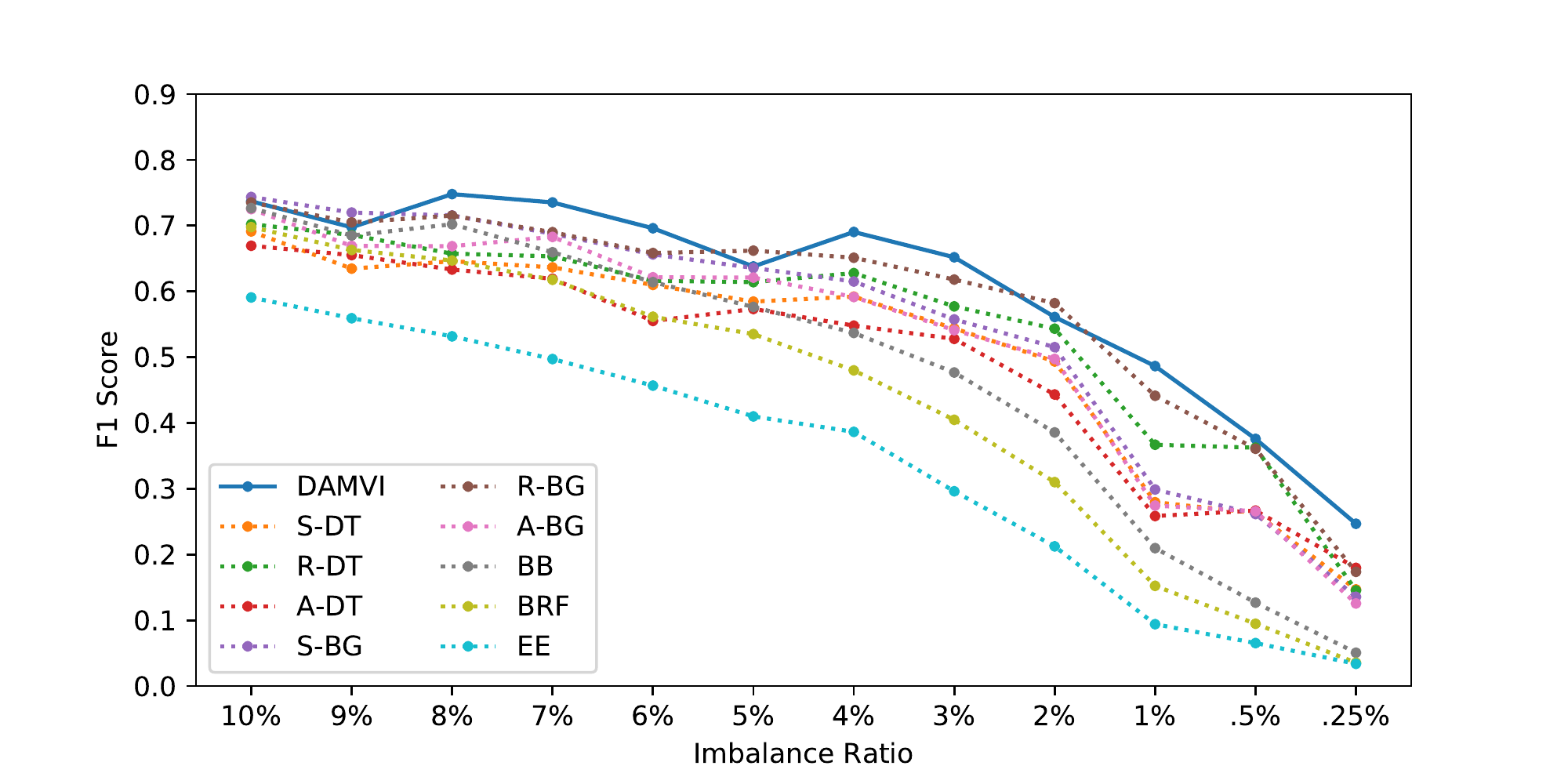}
	\end{subfigure}
	\begin{subfigure}{1.0\textwidth}
		\hspace{-15pt} \includegraphics[scale=0.5]{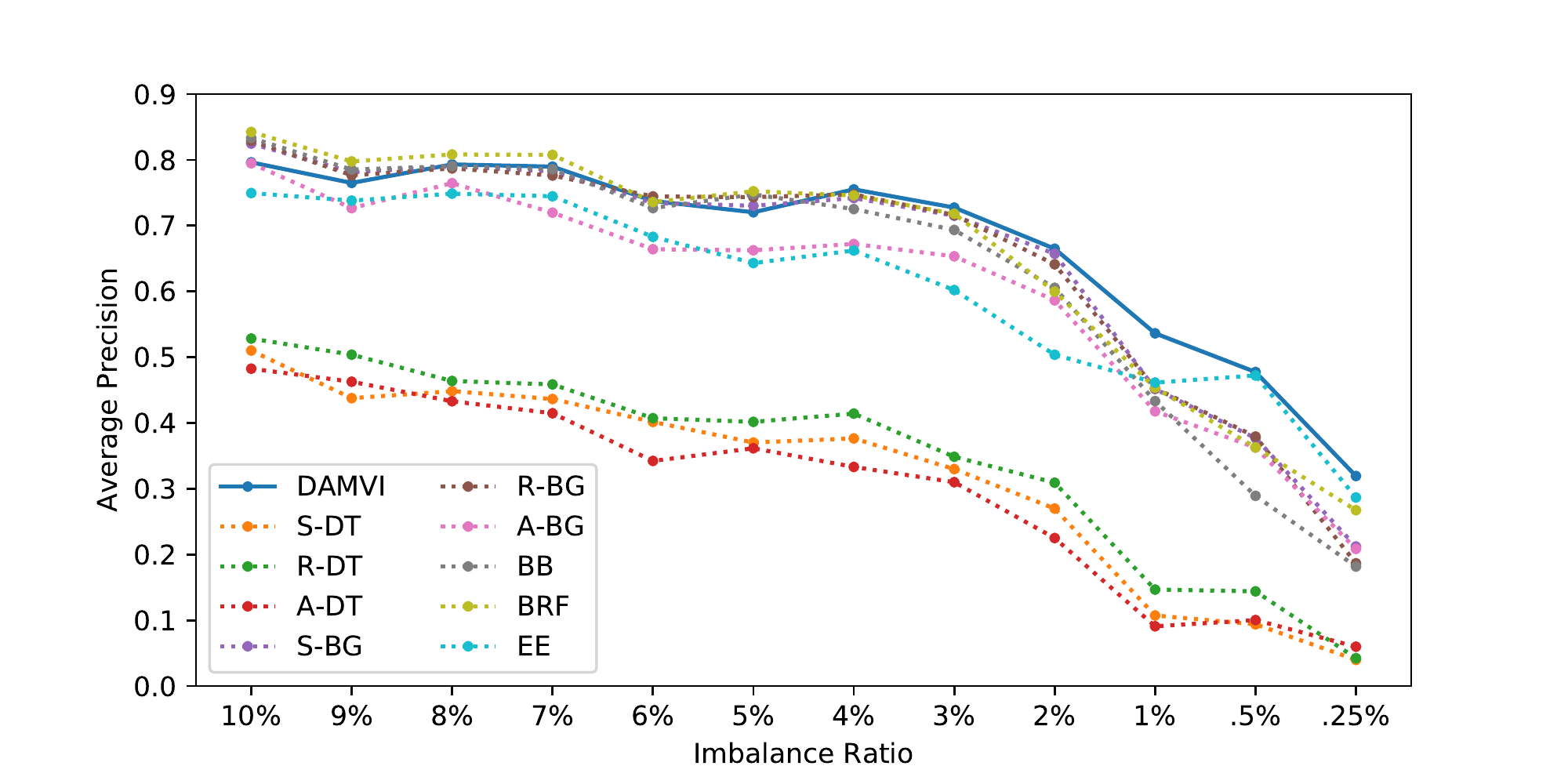}
	\end{subfigure}	
	\caption{Evolution of F1-score and Average Precision {\it w.r.t} Imbalance Ratio on $\cancer$ dataset.} \label{fig:IR_plots}
\end{figure}

We also analyze the behaviour of all the approaches by artificially increasing and decreasing the imbalance for the $\cancer$ dataset.
In order to create a dataset with a higher percentage of minority class examples than in the original dataset, we randomly undersample the majority class examples. Similarly, to create a dataset with a lower percentage of minority class examples than in the original dataset, we randomly undersample the minority class examples. 
Figure \ref{fig:IR_plots} illustrates the obtained results by showing the evolution of F1-score and Average Precision with respect to the imbalance ratio (i.e. percentage of positive class examples) on the $\cancer$ dataset.
As shown in Figure \ref{fig:IR_plots}, $\our$ performs better than baselines both in terms of F1-score and AP when the imbalance ratio (IR) is less than $4\%$ (except at $2\%$ for F1-score ). 
This shows that $\our$ performs well even for highly imbalanced classification tasks ($<4\%$ of IR). 
Below $1\%$ of IR, we can notice that EasyEnsemble ($\ee$) gradually performs second best in terms of AP (but worst in terms of F1-score) and ROSBagging ($\rbg$) performs second best in terms of F1-score (but   drastically drops in terms of AP). 
However, our approach $\our$ remains ``consistent'' and ``stable'' both in terms of F1-score and AP throughout the evolution of imbalance ratio.
This shows that explicitly controlling the trade-off between the accuracy and the diversity among classifiers (by focusing on ``hard''positive examples) plays an important role while learning an ensemble of classifiers for imbalanced datasets.\\[2mm]
\noindent \textbf{A note on the Complexity of the Algorithm:} The complexity of learning a decision tree classifier is $O(d.n.\log(n))$, where $d$ is the dimension of input space. 
We learn the weights over the base classifiers by optimizing Equation (\ref{eq:Cbound} (Step 8 of our algorithm) using SLSQP method which has time complexity of $O(K\pwr 3)$. Therefore, the overall complexity of $\our$ is $O(K.d.n.\log(n) + K\pwr 3)$ . Note that we can easily parallelize $\our$: by using $K$ machines, we can learn decision tree classifiers parallelly and weights over them.

\section{Conclusion}
\label{sec:conclusion}
In this paper, we considered the problem of imbalanced learning where the number of negative examples (majority class) significantly outnumbers the positive class (minority class or class of interest) examples. 
In order to deal with imbalanced datasets, we propose an ensemble learning based algorithm (referred to as $\our$) that learns a diversity-aware weighted majority vote classifier over the base classifiers. 
After learning base classifiers, the algorithm i) increases the weights of positive examples (minority class) which are ``hard'' to classify with uniformly weighted base classifiers; and ii) then learns weights over base classifiers by optimizing the PAC-Bayesian $C$-Bound.
We have validated our approach on various datasets and we show that $\our$ consistently performs better than state-of-art models. 
We also show that explicitly controlling the trade-off between the accuracy and the diversity among base classifiers (with focus on hard positive examples) is an effective strategy to deal with highly imbalanced datasets.

As future work, we would like to extend our algorithm to the \textit{semi-supervised case}, where one has access to an additionally unlabeled set during the training.
One possible way is to learn base classifiers using pseudo-labels (for unlabeled data) generated from the K-means classifier trained using labeled data. 
We would also like to extend our algorithm to the case of multiclass imbalanced classification problems.
One possible solution is to make use of multiclass $C$-Bound\cite{laviolette2014generalizing} to learn the diversity-aware weighted majority vote classifier.

\section{Appendix}
\subsection{Proof of $C$-Bound}
\label{sec:proof_c_bound}
In this section, we present the proof of $C$-Bound (Equation \ref{eq:Cbound}), similar to the proof provided by Germain \textit{et al.}\cite{GermainLLMR15}. 
Firstly, we need to define the margin of the weighted majority vote $B_Q$ and its first and second statistical moments.
\begin{definition}
	\label{def:margin}
	Let $M_{\post}$ is a random variable that outputs the margin of the weighted  majority vote on the example $(x,y)$ drawn from distribution $\mathcal{D}$, given by:
	$$M_{\post} (x,y) =  \E{h \sim \posterior} y\, h(x). $$
	The first and second statistical moments of the  margin are respectively given by
	\begin{equation}
	\label{eq:FirstMoment}
	\mu_{1}(M_{\posterior}^{\mathcal{D}}) = \E{(x,y) \sim \mathcal{D}} M_{\posterior} (x,y). 
	\end{equation}
	and, 
	\begin{align}
	\label{eq:SecondMoment}
	\nonumber \mu_2 (M_{\posterior}^{\mathcal{D}}) & =  \E{(x,y) \sim \mathcal{D}} \big[ M_{\posterior} (x,y) \big]^2 \\
	&= \E{(x,y) \sim \mathcal{D}}    y^2\,\Big[\E{h \sim \posterior} h(x^v) \Big]^2 =\E{(x,y) \sim \mathcal{D}} \Big[  \E{h \sim \posterior} h(x) \Big]^2.
	\end{align}
\end{definition}
\noindent According to this definition, the risk of the weighted majority vote can be rewritten as follows:
\begin{align*}
R_{\mathcal{D}}(B_{\posterior}) = \Pr_{(x,y) \sim \mathcal{D}}  \big( M_{\posterior} (x,y) \le 0 \big).
\end{align*}
\noindent Moreover, the risk of the Gibbs classifier can be expressed thanks to the first statistical moment of the margin.
Note that in the binary setting where $y\in\{-1,1\}$ and $h:\Xcal\to\{-1,1\}$, we have 
$\I{[h(x) \neq y]} = \frac12 (1-y\,h(x))$, and therefore
\begin{align}
R_{\mathcal{D}}(G_{\posterior}) 
&  \nonumber 
= \E{({x},y) \sim \mathcal{D}}  \ \E{h \sim \posterior} \I{[h(x) \neq y]} \\
& \label{eq:rrr}
= \frac{1}{2}\bigg(1- \E{(x,y) \sim \mathcal{D}}  \E{h \sim \posterior} y\,h(x) \bigg)  \\
& \nonumber
= \frac{1}{2}(1-\mu_{1}(M_{\posterior}^{\mathcal{D}}))\,.
\end{align}
Similarly, the expected disagreement can be expressed thanks to the second statistical moment of the margin by
\begin{align}
d_{\Dcal} (\posterior) 
&  \nonumber
=  \E{(x,y) \sim \mathcal{D}}   \E{h \sim \posterior}  \E{h' \sim \posterior} \I{[ h(x) {\ne} h'(x)]}\\
&  \nonumber 
= \frac{1}{2} \bigg( 1- \E{(x,y) \sim \mathcal{D}}   \E{h \sim \posterior}  \E{h' \sim \posterior}  h(x) \times h'(x)  \bigg) \\
&  \nonumber
= \frac{1}{2} \bigg( 1- \E{(x,y) \sim \mathcal{D}} \Big[   \E{h \sim \posterior}   h(x) \Big] \times \Big[   \E{h' \sim \posterior}  h'(x) \Big] \bigg) \\
& \label{eq:ddd}
= \frac{1}{2} \bigg( 1- \E{(x,y) \sim \mathcal{D}}\bigg[ \E{h \sim \posterior} h(x) \bigg]^2  \bigg) \\
& \nonumber
= \frac{1}{2}(1-\mu_{2}(M_{\posterior}^{\mathcal{D}}))\,.
\end{align}
From above, we can easily deduce that $0 \le d_{\Dcal} (\posterior)  \le 1/2$ as $ 0 \le \mu_{2}(M_{\posterior}^{\mathcal{D}}) \le 1$. Therefore, the variance of the margin can be written as:
\begin{equation}
\label{VarianceMargin}
\begin{split}
\text{Var}(M_{\posterior}^{\mathcal{D}}) & =   \underset{(x,y) \sim \mathcal{D}}{\textbf{Var}} (M_{\posterior} (x,y)) \\
& = \mu_2 (M_{\posterior}^{\mathcal{D}}) - (\mu_1 (M_{\posterior}^{\mathcal{D}}))^2.
\end{split}
\end{equation}

\subsection*{The proof of the  $\mathcal{C}$-bound} 
\begin{proof}
	By making use of one-sided Chebyshev inequality (Theorem \ref{theo:chebyshev} in Appendix \ref{sec:mathematical_tools}), with $X=-M_{\posterior} (x,y)$, $\mu= \E{(x,y) \sim \mathcal{D} } (M_{\posterior} (x,y))$ and $a= \E{(x,y) \sim \mathcal{D}} M_{\posterior} (x,y)$, we have 
	\begin{align*}
	& R_{\mathcal{D}}(B_{\posterior})   = \Pr_{(x,y) \sim \mathcal{D}}  \big( M_{\posterior} (x,y) \le 0 \big) \\
	& = \!\!\!\!\! \Pr_{(x,y) \sim \mathcal{D}} \!\! \bigg ( \!\! -M_{\posterior} (x,y)\! +\!\! \E{(x,y) \sim \mathcal{D}} \! M_{\posterior} (x,y)\!\! \ge\!\! \E{(x,y) \sim \mathcal{D}} M_{\posterior} (x,y) \bigg) \\
	& \le \frac{\underset{(x,y) \sim \mathcal{D}}{\textbf{Var}} (M_{\posterior} (x,y))}{\underset{(x,y) \sim \mathcal{D}}{\textbf{Var}} (M_{\posterior} (x,y)) + \bigg( \E{(x,y) \sim \mathcal{D}} M_{\posterior} (x,y) \bigg)^2} \\
	& = \frac{\text{Var} (M_{\posterior}^{\mathcal{D}})}{\mu_2(M_{\posterior}^{\mathcal{D}}) - \bigg(\mu_1(M_{\posterior}^{\mathcal{D}})\bigg)^2 + \bigg(\mu_1(M_{\posterior}^{\mathcal{D}})\bigg)^2}\\
	& = \frac{\text{Var} (M_{\posterior}^{\mathcal{D}})}{\mu_2(M_{\posterior}^{\mathcal{D}})} \\
	& = \frac{\mu_2(M_{\posterior}^{\mathcal{D}}) - \bigg(\mu_1(M_{\posterior}^{\mathcal{D}})\bigg)^2}{\mu_2(M_{\posterior}^{\mathcal{D}})} \\
	& = 1 - \frac{\bigg(\mu_1(M_{\posterior}^{\mathcal{D}})\bigg)^2}{\mu_2(M_{\posterior}^{\mathcal{D}})} \\
	& = 1 - \frac{\bigg(1-2\,R_{\mathcal{D}}(G_{\posterior}) \bigg)^2}{1-2\,d_{\Dcal} (\posterior) }
	\end{align*}
\end{proof}

\subsection{Mathematical Tools}
\label{sec:mathematical_tools}
\begin{theorem}[Cantelli-Chebyshev inequality]
	\label{theo:chebyshev}
	For any random variable $X$ {\it s.t.} $\mathbb{E}(X)\!=\! \mu$ and $\mathbf{Var}(X)=\sigma^2$, and for any $a\!>\!0$, we  have
	$\mathbb{P}(X - \mu \ge a) \le \frac{\sigma^2}{\sigma^2 + a^2}.$
\end{theorem}

\section*{Acknowledgment}
\begin{floatingfigure}[l]{0.23\linewidth}
	\includegraphics[scale=.018]{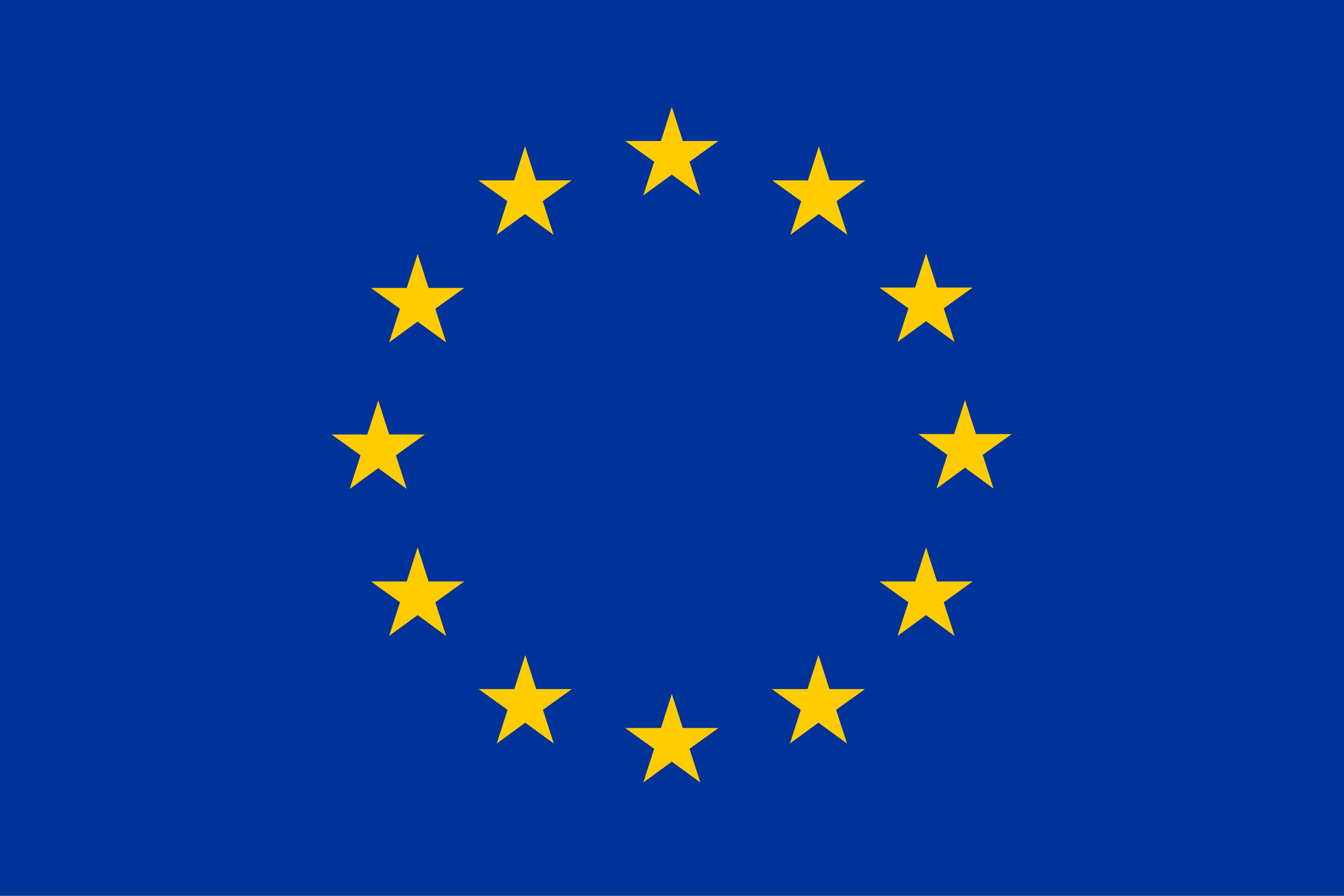} 
\end{floatingfigure}
\noindent This project has partially received funding from the European Union’s Horizon 2020 research and
innovation programme under the grant agreement No 768994. 
The content of this paper does not reflect the official opinion of the European Union.
Responsibility for the information and views expressed therein lies entirely with the author(s).

\bibliographystyle{IEEEtran}
\bibliography{biblio}

\end{document}